\documentclass{article}

\usepackage{amssymb}
\usepackage{subcaption}
\usepackage{mathtools}
\usepackage{graphics}
\usepackage{hyperref}
\usepackage{cite}

\providecommand{\keywords}[1]
{
  \small	
  \textbf{\textit{Keywords---}} #1
}

\title{A Computer Vision-Based Quality Assessment Technique for the automatic control of consumables for analytical laboratories}
\author{Meriam Zribi  \\
	Università degli Studi Roma La Sapienza  \\
    Department of Basic and Applied Sciences for Engineering \\
    Via Antonio Scarpa 14 \\
    Rome
	\and 
	Paolo Pagliuca \\
	National Research Council \\
    Institute of Cognitive Sciences and Technologies \\
    Via Gian Domenico Romagnosi 18/A \\
    Rome
    \and
    Francesca Pitolli  \\
	Università degli Studi Roma La Sapienza  \\
    Department of Basic and Applied Sciences for Engineering \\
    Via Antonio Scarpa 14 \\
    Rome
	}

\date{}

\begin{document}

\maketitle

\begin{abstract}
The rapid growth of the Industry 4.0 paradigm is increasing the pressure to develop effective automated monitoring systems. Artificial Intelligence (AI) is a convenient tool to improve the efficiency of industrial processes while reducing errors and waste. In fact, it allows the use of real-time data to increase the effectiveness of monitoring systems, minimize errors, make the production process more sustainable, and save costs. In this paper, a novel automatic monitoring system is proposed in the context of production process of plastic consumables used in analysis laboratories, with the aim to increase the effectiveness of the control process currently performed by a human operator. In particular, we considered the problem of classifying the presence or absence of a transparent anticoagulant substance inside test tubes. Specifically, a hand-designed deep network model is used and compared with some state-of-the-art models for its ability to categorize different images of vials that can be either filled with the anticoagulant or empty. Collected results indicate that the proposed approach is competitive with state-of-the-art models in terms of accuracy. Furthermore, we increased the complexity of the task by training the models on the ability to discriminate not only the presence or absence of the anticoagulant inside the vial, but also the size of the test tube. The analysis performed in the latter scenario confirms the competitiveness of our approach. Moreover, our model is remarkably superior in terms of its generalization ability and requires significantly fewer resources. These results suggest the possibility of successfully implementing such a model in the production process of a plastic consumables company.
\end{abstract}


\section{Introduction}
\label{intro}

The pressures on the process industry to improve yields, reduce waste and increase profits make it essential to increase the efficiency of process operations \cite{ghobakhloo2020industry}.
In this respect, monitoring systems play a crucial role in making production processes efficient \cite{reis2017industrial}. Unfortunately, traditional monitoring systems are increasingly unable to meet the highly dynamic and demanding requirements of today's industrial applications since they require the presence of various types of sensors (e.g., acoustic, electrical, magnetic, optical, thermal, ultrasonic or visual) \cite{liang2004machining,scott2018industrial} or service personnel. However, sensors are highly sensitive to noise perturbations and their accuracy is highly dependent on their calibration and the environmental conditions. In addition, the evaluation of human operators is subject to their expertise and physical (e.g., fatigue) and mental (e.g., stress) conditions.

One way to overcome these issues is to adopt the Industry 4.0 paradigm \cite{kagermann2011industrie,kagermann2013recommendations,popkova2019industry}, which is growing exponentially \cite{ghobakhloo2020industry}. Industry 4.0 means redefining processes, equipment and technologies in order to address challenges like sustainability, innovation, digital transformation, technological components and skills of the workers \cite{mohamed2018challenges}. Specifically, Industry 4.0 requires companies to use smart sensors (``Sensor 4.0'' \cite{javaid2021significance,porokhnya2023role,schutze2018sensors}) and train skilled operators (``Operator 4.0'' \cite{romero2016towards,romero2020operator,segura2020visual}). Successful examples of adopting the Industry 4.0 paradigm can be found in supply chain management \cite{fatorachian2021impact,ghadge2020impact}, in the chemical industry \cite{he2023review,vaccari2021optimally} and in the food industry \cite{hassoun2023fourth}, although the lack of a common vision among companies still limits the diffusion of this approach \cite{khin2022factors,suleiman2022industry}. 

To achieve the goals of Industry 4.0, one of the key elements is Artificial Intelligence (AI) \cite{jan2023artificial,uraikul2007artificial,zhao2022perspectives}. Artificial Intelligence is indeed considered one of the key technologies for creating autonomous, collaborative production units with advanced capabilities for self-optimization and self-monitoring \cite{kagermann2013recommendations,monostori2014cyber,vogel2014coupling,wu2020concept}. The aim is to use real-time data, analytics and AI techniques to improve production efficiency, flexibility and agility, and to develop new business models based on customized and sustainable production \cite{jan2023artificial,peres2020industrial}. AI might also help mitigate pollution, a heavy social and environmental concern \cite{chan2003artificial,yadav2022reduction}. In fact, one possible countermeasure to reduce waste and preserve ecosystems is process automation \cite{cheremisinoff2013waste}. Therefore, the use of AI approaches in industrial processes might play a key role. In \cite{ahmed2023deep,huynh2020automated,zhang2021recyclable} Convolutional Neural Network (CNN) models are employed to classify and automatically sort waste. Similar approaches exploting deep learning for waste identification are those reported in \cite{narayan2021deepwaste,ramsurrun2021recyclable}. In \cite{panwar2020aquavision} the authors used deep transfer learning to detect waste in water.

The implementation of AI monitoring in the production process can dramatically reduce the waste of non-compliant components, enabling operators to intervene early during the manufacturing process to correct faulty operations. Ultimately, the goal is to increase the accuracy of the quality control process and support the final decision making of the human expert, thereby reducing the cost of the process in terms of both wasted materials and time.

Among the AI techniques, Deep Learning (DL) \cite{lecun1995convolutional} represents a widely used technique in process monitoring \cite{nasir2021review,zhao2019deep}. There are several examples in which deep networks have been successfully employed in such scenarios \cite{iqbal2019fault,lee2017convolutional,liao2021manufacturing,lyu2018image,mu2021industrial,wu2020self,yu2019active}. In \cite{he2021efficient} the authors used a deep CNN model to solve the problem of detecting smoke in normal and foggy environments. Specifically, they employed the VGG-16 deep network \cite{simonyan2014very} combined with an attention mechanism attributing more importance to parts of the image containing smoke elements. In addition, they introduced a feature-level and decision-level module in order to extract more discriminative features enabling to distinguish between smoke and similar objects. A similar work is reported in \cite{khan2019energy}. In \cite{villalba2019deep}, a deep CNN is applied to address the problem of producing gravure cylinders in the printing industry. 

In this paper, we present an automated AI system for monitoring the content of plastic vials, which was designed to meet a medical company's need to certify the quality of in vitro diagnostic devices. Currently, a human operator is responsible for this control, which consists of checking for the presence of a transparent anticoagulant in the test tubes. In addition, the inspection is limited to the external vials of each batch of the tubes produced: if the operator detects a defective vial, the entire batch is discarded, with inevitable disposal costs and high environmental impact. Overall, this control is not efficient and leads to numerous errors that make the tubes unusable. We propose an innovative monitoring system using Computer Vision (CV) to make the process efficient. CV systems are faster, more objective and can operate continuously to inspect thousands of vials per minute. They also provide more accurate inspection results than human operators when they are fine-tuned. By using CV systems, manufacturers can save costs and increase profits by reducing errors, improving yields and ensuring regulatory compliance \cite{javaid2022exploring}.

The key idea behind the development of the proposed automatic certification system is to use a camera that allows the process to be monitored and certified by acquiring images at the critical steps of its execution. Specifically, we developed a new deep CNN model for the detection of the anticoagulant in the vials of an in vitro diagnostic device during the manufacturing process. We trained the proposed model for two different scenarios. In the first, the CNN is asked to detect the presence or the absence of the anticoagulant inside the vials, while in the second the model has to classify not only the presence or the absence of the anticoagulant but also the size of the vial. We compared our CNN model with state-of-the-art deep neural network architectures. Results of our analysis clearly show how the proposed approach is competitive with best models in the first scenario, while it is superior in the most complex scenario. Moreover, our model displays a better generalization capability, which turns out to be crucial for the development of efficient automatic monitoring systems. To our knowledge, the detection of nebulized chemicals within plastic laboratory instrumentation has not been previously addressed in the literature. 

The remainder of the article is organized as follows: Section~\ref{system} contains a description of the considered monitoring system. The proposed model is introduced in Section~\ref{method}. The experimental settings, the results we achieved and their analysis are presented in Section~\ref{results}. Finally, in Section~\ref{conclusions} we provide our conclusions and final remarks.

\section{The Monitoring System}
\label{system}

\begin{figure}[htbp]
    \centering
    \begin{subfigure}{0.49\textwidth}
        \centering
        \includegraphics[scale=0.5]{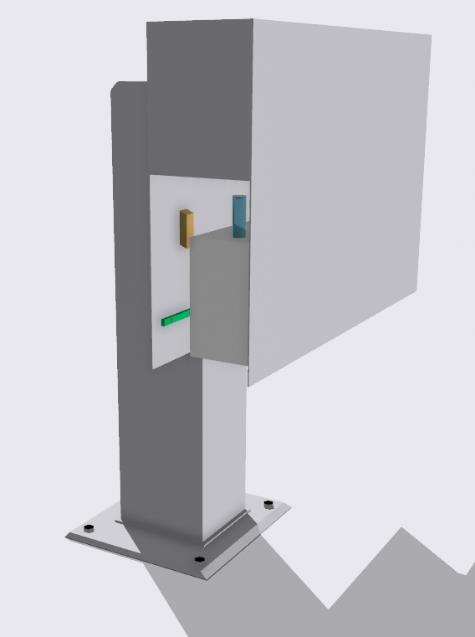}  
        \caption{}
    \end{subfigure}
    \begin{subfigure}{0.49\textwidth}
        \centering
        \includegraphics[scale=0.5]{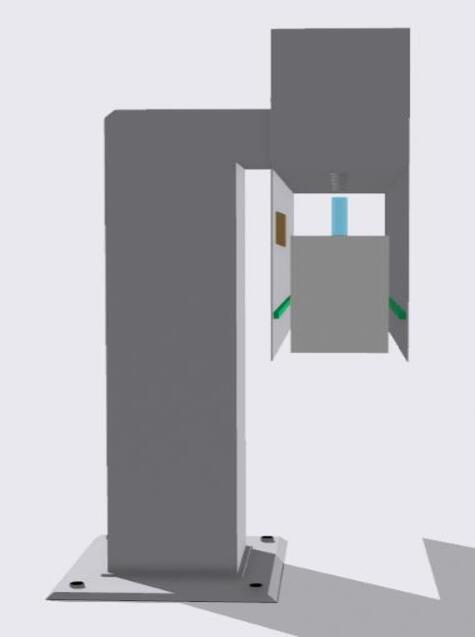}
        \caption{}
    \end{subfigure}
    \caption{The monitoring system. Tubes (blue items) run within the tunnel. The system is illuminated by led cards (green items). A photosensor (orange item) gives the signal for the acquisition of the image.}
    \label{fig:system}
\end{figure}

To make the process efficient and sustainable, the monitoring system must be embedded in the production line of the company performing the test tube analysis. The architecture of the system, shown in Fig.~\ref{fig:system}, consists of a metal frame attached to the floor on one side to give stability to the whole system. At the top of the structure there is a tunnel-shaped element, in which the transparent tubes containing the anticoagulant (Fig.~\ref{fig:system}, blue item) run. The structure is designed to shield the area in which the tubes are analyzed from ambient lighting, which could alter the results. The acquisition sysyem, consisting of the acquisition board and the camera, is located in the false ceiling of the tunnel. The other two elements of interest highlighted in Fig.~\ref{fig:system} are: the led card at the bottom (the green strip) illuminating the system and the photosensor responsible for sending the trigger for the acquisition of images (the orange item).

\begin{figure}[htbp]
    \centering
    \includegraphics[scale=0.35]{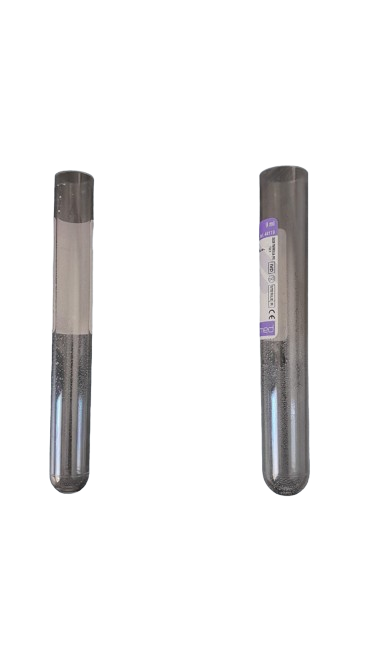}
    \caption{Example of analyzed test tubes. Vials have a label on the side wall. The unit of measurement is the milliliter. \textbf{(left)} a small vial (height: 10 cm, diameter: 1 cm, capacity: 6 ml); \textbf{(right)} a large vial (height: 10 cm, diameter: 1.3 cm, capacity: 9 ml).}
    \label{fig:tubes}
\end{figure}

\begin{figure}[htbp]
    \centering
    \begin{subfigure}{0.45\textwidth}
        \centering
        \includegraphics[scale=0.5]{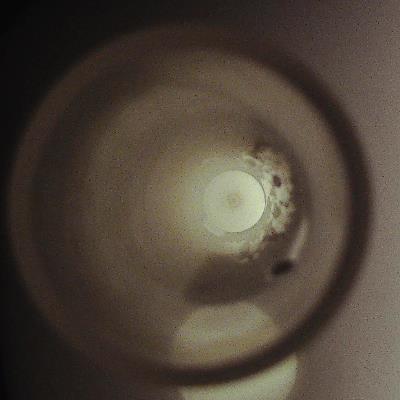}  
        \caption{}
    \end{subfigure}
    \begin{subfigure}{0.45\textwidth}
        \centering
        \includegraphics[scale=0.5]{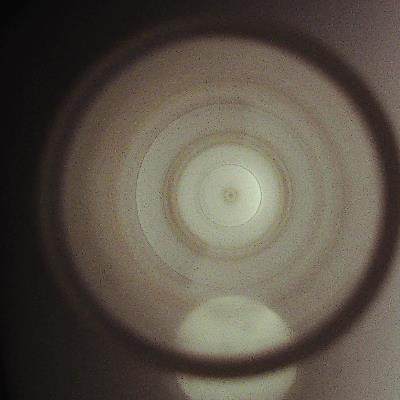}
        \caption{}
    \end{subfigure}
    \caption{Images recorded by the camera. Left figure: a tube containing anticoagulant. Right figure: an empty tube. Figure taken from \cite{zribi2023convolutional}.}
    \label{fig:images}
\end{figure}

Our goal is to detect the presence of the anticoagulant inside the test tubes that are placed inside white racks. The vials can be of different types and sizes and come with a label on the side wall. The anticoagulant may appear either as a droplet at the bottom of the tube, or as a spray on the side walls of the tube. Our tests have shown that the best setup for detecting the anticoagulant is the one in which the camera views the test tube from above since the presence of the label on the side walls limits the size of the scene to be analyzed. An example of the vials to be checked is provided in Fig.~\ref{fig:tubes}. The configuration of the monitoring system we chose allows us to capture images as those shown in Fig.~\ref{fig:images}.

\section{Materials and Methods}
\label{method}

We are interested in two different scenarios that may occur during the manufacturing process:

\begin{enumerate}
    \item detecting the presence or the absence of the anticoagulant regardless of the size of the test tube (referred as $2_{{output-labels}}$ case);
    \item detecting the presence or the absence of the anticoagulant and the size (i.e., large or small) of the test tube (termed as $4_{{output-labels}}$ case).
\end{enumerate} 

To address these challenges, two different strategies have been examined. The first is to create a CNN network from scratch \cite{zribi2023convolutional}, which is illustrated in the next section. The second is the transfer learning technique \cite{torrey2010transfer}, which uses a pre-trained network and re-trains only the last layer with the data from the application of interest. We present the pre-trained deep network models we used in Section~\ref{learning}.

\subsection{The CNN model}
\label{model}

In this section we describe the CNN model we designed, which proved to be effective in a preliminary study we conducted \cite{zribi2023convolutional}. We will refer to this model as \textbf{ConvNet3\_4} since, as we will illustrate later, it consists of 3 convolutional layers and 4 fully-connected layers.

\begin{figure}[htbp]
  \begin{center}
    \includegraphics[angle=0,width=0.5\textwidth]{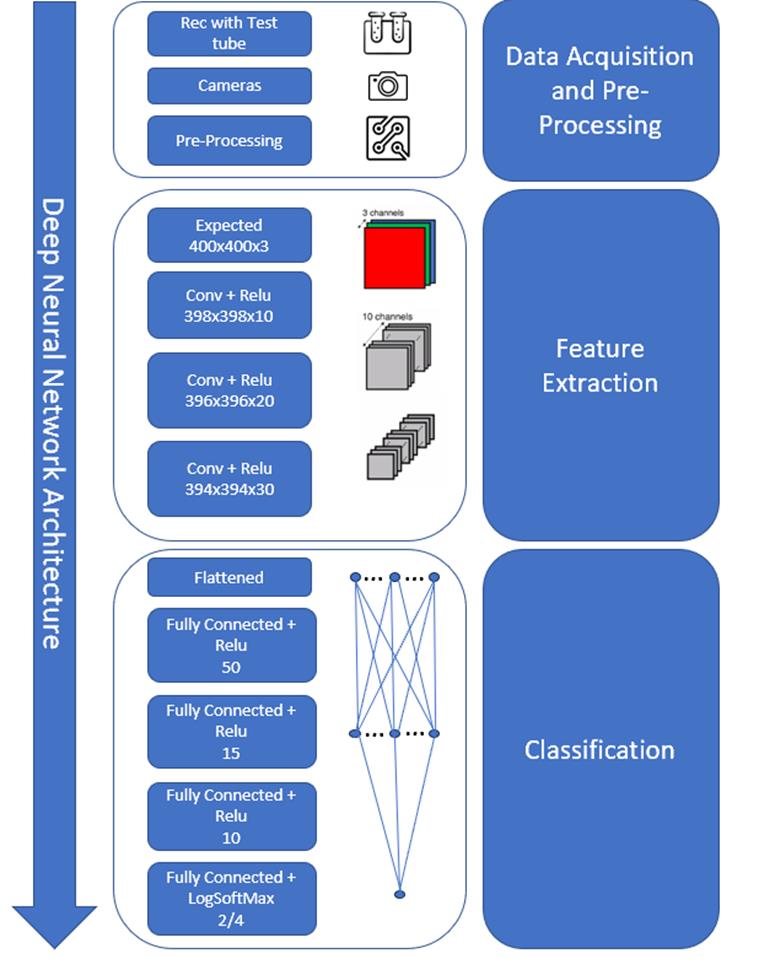}
    \caption{Deep Neural Network Architecture.}
    \label{fig:model}
  \end{center}
\end{figure}

\begin{figure}[htbp]
  \begin{center}
    \includegraphics[angle=0,width=\textwidth]{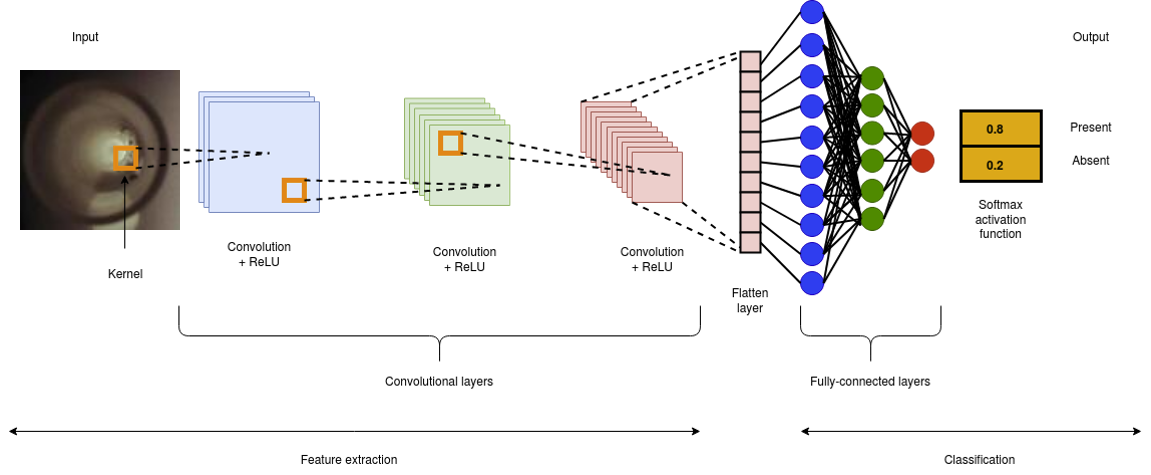}
    \caption{Illustration of the ConvNet3\_4 model.}
    \label{fig:convNet}
  \end{center}
\end{figure}

The whole monitoring process works as illustrated in Fig.~\ref{fig:model}, while the ConvNet3\_4 model is shown in Fig.~\ref{fig:convNet}. The model is organized in two main blocks:

\begin{enumerate}
    \item \textbf{Feature extraction} block (\textit{FEB}) consisting of a sequence of Convolutional Layers (CL);
    \item \textbf{Classification} block (\textit{CB}): it is 
    a multi-layer fully-connected network that associates output labels to input data. The input of the classification block corresponds to the output of the feature extraction block.
\end{enumerate}

\subsubsection{Feature Extraction Block (FEB)}

The FEB is responsible for extracting relevant features from input images by convolution with a kernel function. To provide an explanation of the whole process, let us consider an input image represented as a tensor $I$ of size $m_1 \times m_2 \times m_c$, where $m_1$ denotes the height of the image, $m_2$ indicates the width of the image and $m_c$ represents the number of channels of the image. Suppose a filter (the kernel) is applied, which is described by a tensor $K$ of size $n_1 \times n_2 \times n_c$, with $n_c$ equal to $m_c$. The filter $K$ is convolved with the image $I$ resulting in a matrix $F$ of size $(m_1 - n_1 + 1) \times (m_2 - n_2 + 1) \times 1$:

\begin{equation} \label{eq:featureExtraction}
F[i,j] = {(I*K)}_{[i,j]}
\end{equation}

Usually, a bias term $b$ is added to the convoluted part $F$ before an activation function $\phi$ is applied:

\begin{equation}\label{eq:convolutionOutput}
Conv(I,K) = \phi((I*K) + b)
\end{equation}

This process is iterated for the number of convolutional layers in the model:

\begin{equation}\label{eq:featureExtractionProcess}
F_l = Conv(F_{l-1},K) \qquad  l = 2,...,N_{conv\_layers}
\end{equation}

where $N_{conv\_layers}$ denotes the number of convolutional layers in the model and $I = F_1$.

Although not included in the ConvNet3\_4 model, in many cases a convolutional layer is followed by a pooling layer, which reduces the spatial size of the convoluted features and extracts the most relevant one \cite{goodfellow2016deep}. However, we decided to avoid the use of pooling layers since experimental results (not shown) demonstrated that the technique does not provide any advantage in terms of accuracy. Conversely, the generalization capability of the model decreases. This implies that pooling has a negative impact on the performance of the model in the considered scenario.

Once the feature extraction is completed, the resultant $F_{N_{conv\_layers}}$ is flattened into a single vector $x$ that will be the input of the CB.

\subsubsection{Classification Block (CB)}

The CB consists in a sequence of one or more fully-connected layers, each of which performs the following computation:

\begin{equation}\label{eq:fcComputation}
\begin{split}
X = w \times x + b \\ z = \gamma(X)
\end{split}
\end{equation}

where $\gamma$ is an activation function for the fully-connected layer. Also in this case, the process is repeated for the number of fully-connected layers in the model:

\begin{equation}\label{eq:classificationProcess}
\begin{split}
X_l = w \times z_{l-1} + b \qquad  l = 2,...,N_{fc\_layers} \\
z_l = \gamma(X_l) \qquad  l = 2,...,N_{fc\_layers}
\end{split}
\end{equation}

where $N_{fc\_layers}$ is the number of fully-connected layers. Once the process is terminated, the output of the last fully-connected layer indicates the probabilities that the input image belongs to each of the possible classes. In order to select among these options, the output passes through a function $\psi$ (e.g., softmax as in the ConvNet3\_4 model or linear regression) to extract the more likely option, which represents the final model's output:

\begin{equation}\label{eq:modelOutput}
\hat{y} = \psi(z_{N_{fc\_layers}})
\end{equation}

\subsubsection{Backpropagation algorithm}

The whole process we have just described allows a deep convolutional network like the ConvNet3\_4 model to predict an output $\hat{y}$ for a given input image. However, in order to make the model learn the association between the input data (i.e., images) and the outputs (i.e., the labels), a backward process is required to adjust the model parameters so as to minimize the error between actual outputs and predictions. This is typically made by using the backpropagation algorithm \cite{rumelhart1986learning}. We now provide an explanation of the method. Suppose we have a feed-forward network to be trained, which should associate the expected output $y$ to a given input $x$. The network is usually initialized with random weights. Now, suppose the model produces the output $\hat{y}$, which is different from the expected value $y$. We can compute the output error $E = E(y,\hat{y})$ as a function of $y$ and $\hat{y}$. Examples of error functions can be the Mean Squared Error or the Cross-Entropy Loss (the one used in this work). The backpropagation algorithm acts by adjusting the weights and biases of the neural network model in order to minimize the output error $E$. The backward process propagates from the output layer up to the first layer of the neural network model (CB in our case). The backpropagation algorithm works also for convolutional layers (FEB in the ConvNet3\_4 model). For further details, see \cite{zhou2018understanding}.

To summarize, the proposed ConvNet3\_4 model first performs a forward pass to extract the relevant features from the input image and predicts the corresponding output (i.e., a label). Then, the error between the expected output and the model prediction is computed and a backward pass is run in order to adjust the model weights and attempt to minimize the error. 

\section{Results}
\label{results}

In this section we describe the parameters of our model (Section~\ref{train_params}) and the different pre-trained networks (Section~\ref{learning}) we used to perform our experiments, we illustrate the dataset used for training (Section~\ref{dataset}) and we show the outcomes of our analyses (Sections~\ref{cross} -~\ref{features}).

\subsection{Training phase}
\label{train_params}

\begin{table}[htbp!]
  \begin{center}
    {\caption{Parameters of the FEB. For each convolutional layer (identified with the layer id), we report the number of input channels, the number of output channels, the kernel size and the stride. We also provide the size of input and output.}\label{tab:paramsCL}}
    \resizebox{\columnwidth}{!}{%
    \begin{tabular}{|c|c|c|c|c|c|c|}
      \hline
      \textbf{\begin{tabular}{@{}c@{}} Layer \\ id \end{tabular}} & \textbf{\begin{tabular}{@{}c@{}} \# input \\ channels \end{tabular}} & \textbf{\begin{tabular}{@{}c@{}} \# output \\ channels\end{tabular}} & \textbf{\begin{tabular}{@{}c@{}} Kernel \\size\end{tabular}} & \textbf{Stride} & \textbf{\# inputs} & \textbf{\# outputs} \\
      \hline
      1 & 3 & 10 & 3 & 1 & $400 \times 400 \times 3$ & $398 \times 398 \times 10$ \\
      \hline
      2 & 10 & 20 & 3 & 1 & $398 \times 398 \times 10$ & $396 \times 396 \times 20$ \\
      \hline
      3 & 20 & 30 & 3 & 1 & $396 \times 396 \times 20$ & $394 \times 394 \times 30$ \\
      \hline
    \end{tabular}%
    }
  \end{center}
\end{table}

The ConvNet3\_4 model has been trained on a dataset of RGB images having a resolution of $400 \times 400$ pixels (see Section~\ref{dataset} for further details). Two examples of the input images are shown in Fig.~\ref{fig:images}. 

The FEB consists of 3 convolutional layers: the first layer has a number of input channels corresponding to the RGB colors. For the other layers, we set:

\begin{equation} \label{eq:convLayerInputs}
n_{in-channels}(l) = n_{out-channels}(l - 1) \qquad l \geq 1
\end{equation}

where \textit{l} is the layer identification number. The parameters of the FEB are listed in Table~\ref{tab:paramsCL}. The outputs of the last convolutional layer are used as inputs for the CB. It consists of 4 fully-connected layers (with identifiers numbered from 4 to 7), whose parameters are listed in Table~\ref{tab:paramsFC}. Last layer generates ${<n>}_{{output-labels}}$ values, each one corresponding to a specific class. In order to select among the different outputs, we applied a softmax layer.

\begin{table}[htbp!]
  \begin{center}
    {\caption{Parameters of the CB. For each fully-connected layer (identified with the layer id), we report the number of inputs and the number of outputs.}\label{tab:paramsFC}}
    \begin{tabular}{|c|c|c|}
      \hline
      \textbf{\begin{tabular}{@{}c@{}} Layer \\ id \end{tabular}} & \textbf{\# of inputs} & \textbf{\# of outputs} \\
      \hline
      4 & $394 \times 394 \times 30$ & 50 \\
      \hline
      5 & 50 & 15 \\
      \hline
      6 & 15 & 10 \\
      \hline
      7 & 10 & ${<n>}_{{output-labels}}$ \\
      \hline
    \end{tabular}
  \end{center}
\end{table}

In this work, we employed the Adam \cite{kingma2014adam} optimizer with weight decay (set to $10^{-2}$), i.e. a regularization technique penalizing large weights in order to keep the size of the network weights within limits. We chose the Adam optimizer since it represents a state-of-the-art optimizer in several DL domains \cite{liquet2024mathematical,reyad2023modified,yaqub2020state}. Both the ConvNet3\_4 and the deep networks described in Section~\ref{learning} have been trained by using the parameters reported in Table~\ref{tab:paramsTrain}. We used the same parameters for both the $2_{{output-labels}}$ and the $4_{{output-labels}}$ cases.

\begin{table}[htbp!]
  \begin{center}
    {\caption{List of parameters with values used for model's training. The parameter \textit{\# of replications} refers only to the ConvNet3\_4 model.}\label{tab:paramsTrain}}
    \begin{tabular}{|c|c|}
      \hline
      \textbf{Parameter} & \textbf{Value} \\
      \hline
      \# of replications & 10 \\
      \hline
      \# of epochs & 50 \\
      \hline
      learning rate & $10^{-4}$ \\
      \hline
      batch size & 64 \\
      \hline
    \end{tabular}
  \end{center}
\end{table}

\subsection{Transfer learning}
\label{learning}

Transfer learning is defined as ``the improvement of learning in a new task through the transfer of knowledge from a related task that has already been learned'' \cite{torrey2010transfer}. Transfer learning is appealing since it allows to train a pre-trained model on new data without starting from scratch. Transfer learning has been successfully used in image classification. In \cite{shaha2018transfer} the authors applied transfer learning to the AlexNet \cite{krizhevsky2012imagenet}, VGG-16 \cite{simonyan2014very} and VGG-19 \cite{simonyan2014very} networks. In \cite{hussain2019study} transfer learning is employed on the Inception v3 model \cite{szegedy2016rethinking}. In \cite{kensert2019transfer} the authors used pre-trained ResNet \cite{he2016deep} and Inception networks to predict cell morphological changes due to chemical perturbations.
\\
In our case, in the absence of similar applications, we studied the problem by assimilating it to more common applications in the literature, such as the detection of fog or smoke \cite{he2021efficient,khan2019energy}. We considered the following models:

\begin{itemize}
    \item VGG-16 \cite{simonyan2014very}
    \item VGG-19 \cite{simonyan2014very}
    \item ResNet-50 \cite{he2016deep}
    \item ResNet-152 \cite{he2016deep}
    \item DenseNet-121 \cite{huang2017densely}
    \item DenseNet-201 \cite{huang2017densely}
    \item AlexNet \cite{krizhevsky2012imagenet}
    \item SqueezeNet1.1 \cite{iandola2016squeezenet}
\end{itemize}

Similarly to other applications using transfer learning techniques, the deep network models we used have been pre-trained on ImageNet \cite{deng2009imagenet} data set for image classification ($14 \times {10}^6$ images) in 1000 different categories. We then refined the original model, adjusting the last fully-connected layer from 1000 outputs to $2_{{output-labels}}$ / $4_{{output-labels}}$ classes to perform the expected classification regarding the presence or the absence of the anticoagulant and/or the type of the test tube.

\subsection{Dataset}
\label{dataset}

Thanks to the setup created in the laboratory, the training and test datasets were acquired by simulating an environment whose lighting was controlled (see Section~\ref{system}). Under these conditions, 402 images were acquired (see the examples in Fig.~\ref{fig:images}). The dataset is composed as follows:

\begin{enumerate}
    \item 101 images with large vial filled with anticoagulant
    \item 100 images with large vial without the anticoagulant
    \item 102 images with small vial full of anticoagulant
    \item 99 images with small vial without the anticoagulant
\end{enumerate}

\begin{table}[htbp!]
  \begin{center}
    {\caption{List of transformations applied to increase the size of datasets used for training and test. Each transformation is applied to each image in the original dataset.}\label{tab:trainTrans}}
    \begin{tabular}{|c|c|c|}
      \hline
      \textbf{Type} & \textbf{\# of transforms} & \textbf{Parameter} \\
      \hline
      Rotation & 5 & random angle in range $[30^\circ,150^\circ]$ \\
      \hline
      Blur & 3 & \begin{tabular}{@{}c@{}} random sigma in range $[0.1,5.0]$ \\ kernel size = 5 \end{tabular} \\
      \hline
      Blur & 3 & \begin{tabular}{@{}c@{}} random sigma in range $[0.1,5.0]$ \\ kernel size = 9 \end{tabular} \\
      \hline
      Posterization & 1 & 2 bits \\
      \hline
      Posterization & 1 & 4 bits \\
      \hline
      Posterization & 1 & 8 bits \\
      \hline
      Sharpness & 1 & sharpness\_factor = 2 \\
      \hline
      Inversion & 1 & \\
      \hline
      Solarization & 1 & threshold = 63 \\
      \hline
      Solarization & 1 & threshold = 127 \\
      \hline
      Solarization & 1 & threshold = 192 \\
      \hline
      Equalization & 1 & \\
      \hline
      Horizontal flipping & 1 & \\
      \hline
      Vertical flipping & 1 & \\
      \hline
    \end{tabular}
  \end{center}
\end{table}

All the images have a resolution of $400 \times 400$ pixels. The dataset is balanced w.r.t. to the different output classes. Nevertheless, due to its extremely small size, the data augmentation technique \cite{tanner1987calculation} was applied. Data augmentation is crucial in DL processes, especially when only few data are available \cite{mikolajczyk2018data,shorten2019survey,taylor2018improving,wang2017effectiveness,yang2022image}. In fact, models trained on small datasets typically suffer from the overfitting issue \cite{shorten2019survey,ying2019overview}, i.e. they are effective at categorizing those data, but unable to generalize to new unseen inputs. Generalization is a paramount property that modern industrial processes must possess. Data augmentation has been largely used in image classification on available datasets, such as CIFAR-10 \cite{gu2019improve,lei2019preliminary} and MNIST \cite{wang2017effectiveness}. In \cite{hernandez2018further} the authors highlighted the benefit of applying data augmentation on CNNs compared to using regularization techniques, like weight decay \cite{krogh1991simple} and dropout \cite{srivastava2014dropout}.

The list of transformations that have been applied to the original dataset is provided in Table~\ref{tab:trainTrans}. Overall, we applied 22 transformations to each image acquired in the laboratory. The final dataset is then composed of 9246 images. The training set consists of 7843 images (85\%), while the test set includes 1403 images (15\%). For more details about each transformation, its effect and the explanation of the parameters, the reader is referred to \cite{pytorch_transforms}.

We want to stress that the dataset used in the two scenarios - $2_{{output-labels}}$ and $4_{{output-labels}}$ - is the same, the only difference is the labeling given to the images w.r.t. the specific experiment. In fact, there are two classifier outputs in the first condition ($2_{{output-labels}}$ case), where the number of outputs is four in the second condition ($4_{{output-labels}}$ case).

\subsection{Cross-validation}
\label{cross}

Since the proposed ConvNet3\_4 model has been designed from scratch, we employed the cross-validation technique \cite{anthony1998cross,schaffer1993selecting} in order to verify whether our deep network is suitable for the considered monitoring problem. Moreover, cross-validation is a widely used tool to avoid overfitting to the input data \cite{arlot2010survey,ghojogh2019theory,yates2023cross}. Specifically, we adopted the $k-fold$ cross-validation method \cite{dietterich1998approximate,guyon2010model}. This technique partitions training data into $k$ disjoint subsets, where the size of each subset is approximately the same: if we denote the number of training data as $N_{tr\_data}$, we can compute the size $N_{tr\_subset}$ of each subset according to the following equation:

\begin{equation}\label{eq:k_subset}
    N_{tr\_subset} = \frac{(k - 1) * N_{tr\_data}}{k}
\end{equation}

\begin{figure}[htbp!]
  \begin{center}
    \includegraphics[angle=0,width=\textwidth]{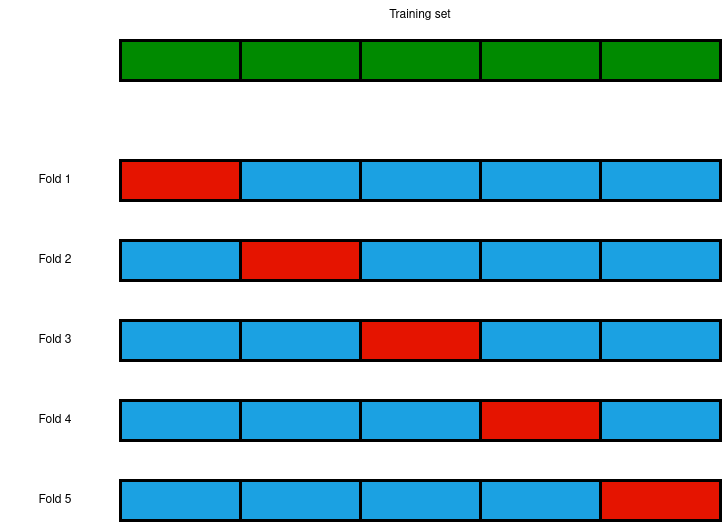}
    \caption{Illustration of the $k-fold$ cross-validation method, with $k = 5$. At each iteration (i.e., the fold), part of the original training data (green in figure) is retained and used as validation set (red in figure). The remaining data (blue in figure) are used as training set for that specific model.}
    \label{fig:kfold}
  \end{center}
\end{figure}

An example of how $k-fold$ cross-validation works is shown in Fig.~\ref{fig:kfold}: it takes the original training data as input and generates $k$ different disjoint subsets, which are then used to evaluate the model.

It is worth noting that, as we stated above, this method allows to verify the suitability of a model for a specific problem. Indeed, at each iteration a new model is initialized and trained. Therefore, the technique returns $k$ different models trained on distinct subsets. The overall accuracy of the model ${acc}_k$ can be computed as:

\begin{equation}\label{eq:kfoldAccuracy}
    {acc}_k = \frac{1}{k} \sum_{i=1}^k {err}_i
\end{equation}

where ${err_i}$ represents the classification error of the $i-th$ fold.

\begin{table}[htbp!]
  \begin{center}
    {\caption{Accuracy of each fold in the $2_{{output-labels}}$ case.}\label{tab:kFoldAccuracy2}}
    \begin{tabular}{|c|c|c|c|c|c|}
      \hline
      \textbf{Fold 1} & \textbf{Fold 2} & \textbf{Fold 3} & \textbf{Fold 4} & \textbf{Fold 5} & \textbf{Average} \\
      \hline
      100\% & 99.946\% & 100\% & 99.946\% & 99.946\% & 99.968\% \\
      \hline
    \end{tabular}
  \end{center}
\end{table}

\begin{table}[htbp!]
  \begin{center}
    {\caption{Accuracy of each fold in the $4_{{output-labels}}$ case.}\label{tab:kFoldAccuracy4}}
    \begin{tabular}{|c|c|c|c|c|c|}
      \hline
      \textbf{Fold 1} & \textbf{Fold 2} & \textbf{Fold 3} & \textbf{Fold 4} & \textbf{Fold 5} & \textbf{Average} \\
      \hline
      99.946\% & 96.322\% & 99.946\% & 100\% & 96.160\% & 98.475\% \\
      \hline
    \end{tabular}
  \end{center}
\end{table}

In this work we set $k = 5$ as in Fig.~\ref{fig:kfold}. Tables~\ref{tab:kFoldAccuracy2} and ~\ref{tab:kFoldAccuracy4} show the accuracy of the 5 different folds for the $2_{{output-labels}}$ and the $4_{{output-labels}}$ cases, respectively. Specifically, cross-validation gives an accuracy score above $98\%$ in both conditions, which is a very good result. This confirms that the proposed ConvNet3\_4 model is suitable for the problem under consideration.

\begin{figure}[htbp!]
    \centering
        \begin{subfigure}[b]{0.45\textwidth}
            \centering
            \includegraphics[width=\textwidth]{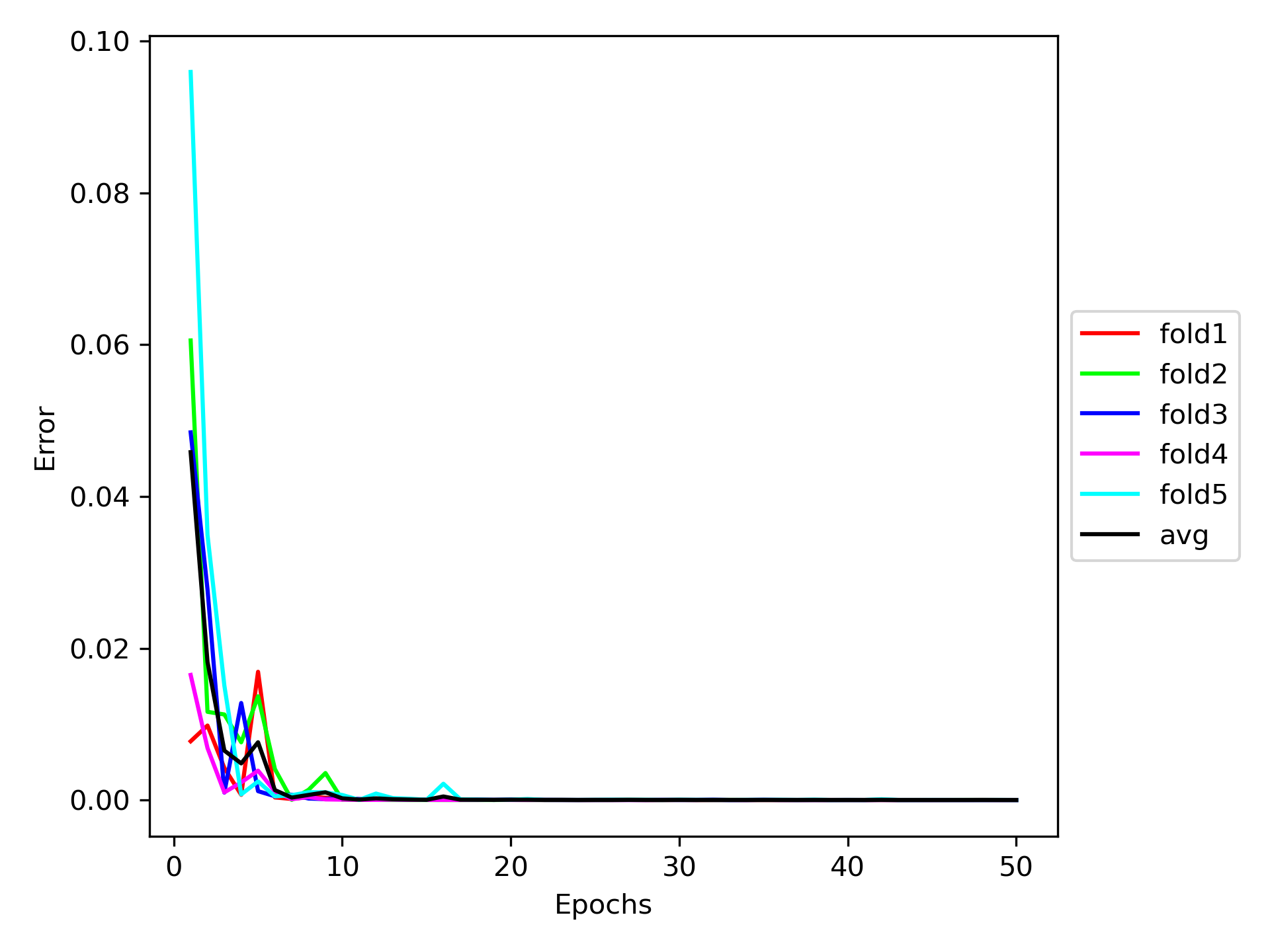}
        \end{subfigure}
        \hfill
        \begin{subfigure}[b]{0.45\textwidth}
            \centering
            \includegraphics[width=\textwidth]{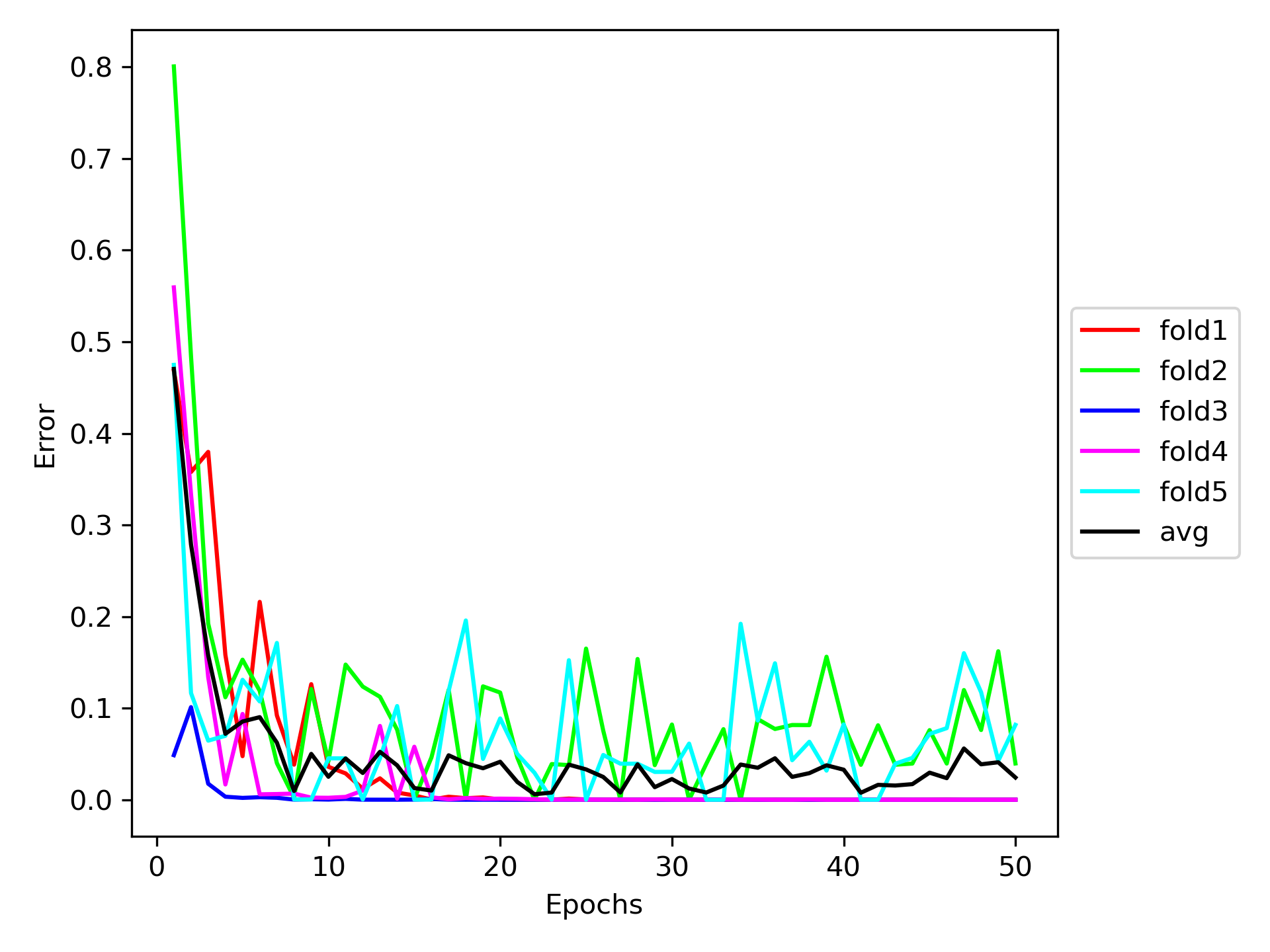}
        \end{subfigure}
    \caption{Training error during the $k-fold$ cross-validation phase. Data refer to the best models returned by the procedure. Left: $2_{{output-labels}}$ case. Right: $4_{{output-labels}}$ case.}
    \label{fig:kFoldModels}
\end{figure}

Fig.~\ref{fig:kFoldModels} shows the error curves when training the $k$ different folds in the $2_{{output-labels}}$ (Fig.~\ref{fig:kFoldModels}, left) and the $4_{{output-labels}}$ cases (Fig.~\ref{fig:kFoldModels}, right). We chose the cross-entropy loss as an error metric. As can be observed, in the first case the error is significantly lower since the classification task is less complex (the error is around 0 after about 10 epochs). In the second case, however, the error curves have different trends depending on the particular fold and oscillations occur. This behavior is not surprising and is strictly dependent on the specific subset each fold has been evaluated with. In particular, because the subsets used in each fold are randomly extracted, it is possible that images are not equally balanced among the different classes.

\subsection{$2_{{output-labels}}$ case}
\label{output2}

In this section we present the results obtained in the $2_{{output-labels}}$ case, which requires to distinguish between presence and absence of the anticoagulant inside vials.

\subsubsection{ConvNet3\_4 model}
\label{convNet3_4-res2}

\begin{figure}[htbp!]
    \centering
        \begin{subfigure}[b]{0.45\textwidth}
            \centering
            \includegraphics[width=\textwidth]{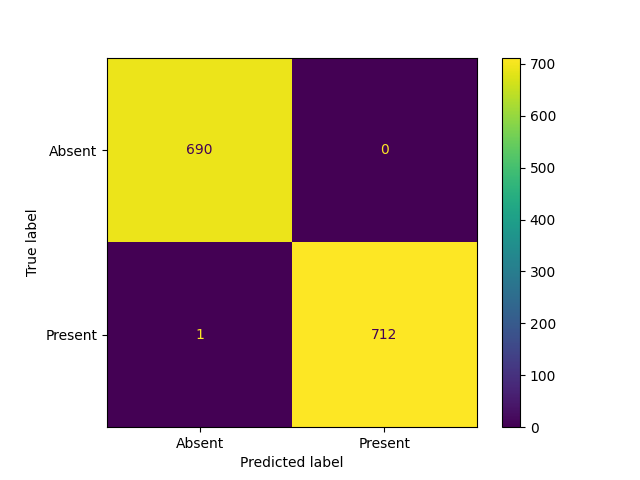}
        \end{subfigure}
        \hfill
        \begin{subfigure}[b]{0.45\textwidth}
            \centering
            \includegraphics[width=\textwidth]{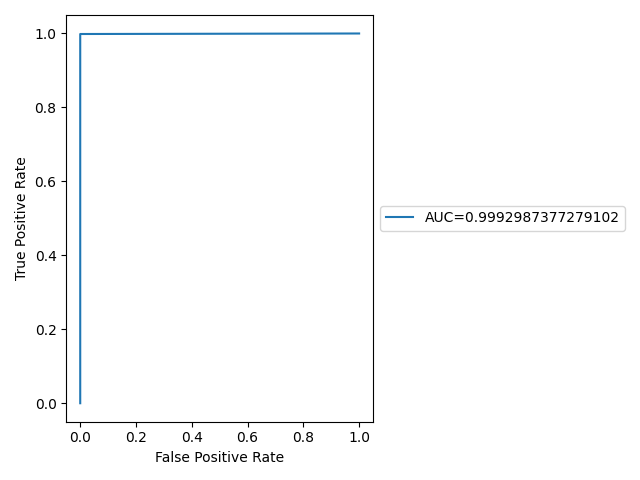}
        \end{subfigure}
    \caption{Classification results of our best model in the $2_{{output-labels}}$ case. Data indicate the classification capability w.r.t. the images in the test set. Left: Confusion matrix. Right: ROC curve.}
    \label{fig:resultsAdam2}
\end{figure}

As we described in Section~\ref{train_params}, we train the ConvNet3\_4 model 10 times, each time starting with different initial weights. This process allows us to minimize the impact of chance related to the extraction of random values. Moreover, repeating the training multiple times permits us to validate thoroughly our approach. At the end of the training process, we obtained an average accuracy of 99.551\% on the test set in the $2_{{output-labels}}$ case, i.e. a quasi-optimal performance. A detailed analysis of the classification capability of the best models trained for the $2_{{output-labels}}$ case is provided in Fig.~\ref{fig:resultsAdam2}. 

As can be observed, our best model successfully categorizes 1402 out of 1403 test images (Fig.~\ref{fig:resultsAdam2}, left). The AUC score is around 0.999 (Fig.~\ref{fig:resultsAdam2}, right), which represents an almost perfect classifier. This outcomes are in agreement with those obtained in \cite{zribi2023convolutional}, although the problem considered here represents a much more challenging task due to both the size of the training set and the complexity of the input images.

\subsubsection{Pre-trained models}
\label{pre-trained-res2}

Table~\ref{tab:accuracy2} contains the outcomes obtained by the different models in the $2_{{output-labels}}$ case. We computed five different metrics: accuracy, recall (or sensitivity), specificity, precision and F1-score.

As stated in Section~\ref{learning}, pre-trained models have been evaluated by using the weights for the training on the ImageNet dataset. To make the comparison fair, we kept the same settings (i.e., \textit{\# of epochs}, \textit{learning rate} and \textit{batch size}) listed in Table~\ref{tab:paramsTrain}. Regarding the ConvNet3\_4 model, we consider the best model out of 10 replications of the experiment.

\begin{table}[htbp!]
  \begin{center}
    {\caption{Performance of the models in the $2_{{output-labels}}$ case. We calculated the following metrics: accuracy (\textit{Acc}), recall (\textit{Rec}), specificity (\textit{Spec}), precision (\textit{Prec}) and F1-score (\textit{F1}). All the metrics are bounded in the range $[0,1]$. As far as the ConvNet3\_4 model is concerned, data refer to the best model out of 10 replications. The bold value indicates the best result we obtained.}\label{tab:accuracy2}}
    \begin{tabular}{|c|c|c|c|c|c|}
      \hline
      \textbf{Model} & \textbf{Acc} & \textbf{Rec} & \textbf{Spec} & \textbf{Prec} & \textbf{F1} \\
      \hline
      ConvNet3\_4 & \textbf{0.9993} & 0.9986 & \textbf{1.0} & \textbf{1.0} & \textbf{0.9993} \\
      \hline
      VGG-16 & 0.9979 & 0.9986 & 0.9971 & 0.9972 & 0.9979 \\
      \hline
      VGG-19 & 0.9979 & 0.9986 & 0.9971 & 0.9972 & 0.9979 \\
      \hline
      ResNet-50 & 0.9765 & 0.9804 & 0.9725 & 0.9735 & 0.9769 \\
      \hline
      ResNet-152 & 0.98 & 0.9762 & 0.9841 & 0.9844 & 0.9803 \\
      \hline
      DenseNet-121 & 0.9979 & 0.9986 & 0.9971 & 0.9972 & 0.9979 \\
      \hline
      DenseNet-201 & 0.9979 & \textbf{1.0} & 0.9957 & 0.9958 & 0.9979 \\
      \hline
      AlexNet & 0.9943 & 0.9986 & 0.9899 & 0.9903 & 0.9944 \\
      \hline
      SqueezeNet1.1 & 0.4918 & 0.0 & 0.4918 & 0.0 & 0.0 \\
      \hline
    \end{tabular}
  \end{center}
\end{table}

As can be seen, all the models manage to achieve excellent results in terms of all the metrics with the exception of the SqueezeNet1.1 model, which classifies less than half of the images (\textit{Acc} = 0.4918). The sub-optimal performance of the latter model can be explained by considering that it classifies all the images according to only one of the two possible outputs (i.e., presence and absence of the anticoagulant).

To verify the capability of the trained models to generalize to new data, we run a post-validation phase on 4 different validation sets of increasing complexity. Some of them contain more than a single transformation. Specifically, for each validation set, we created 200 images for validation, obtained by applying 10 transformations to 20 images in the original dataset. We chose $20 \div {<n>}_{{output-labels}} $ images for each output class to create a uniformly distributed validation set as follows:

\begin{enumerate}
    \item ${val\_set}_1$: obtained by applying the transformations listed in Table~\ref{tab:valTrans1}
    \item ${val\_set}_2$: obtained by applying the transformations listed in Table~\ref{tab:valTrans2}
    \item ${val\_set}_3$: obtained by applying the transformations listed in Table~\ref{tab:valTrans3}
    \item ${val\_set}_4$: obtained by applying the transformations listed in Table~\ref{tab:valTrans4}
\end{enumerate}

\begin{figure}[htbp!]
  \begin{center}
    \includegraphics[angle=0,width=\textwidth]{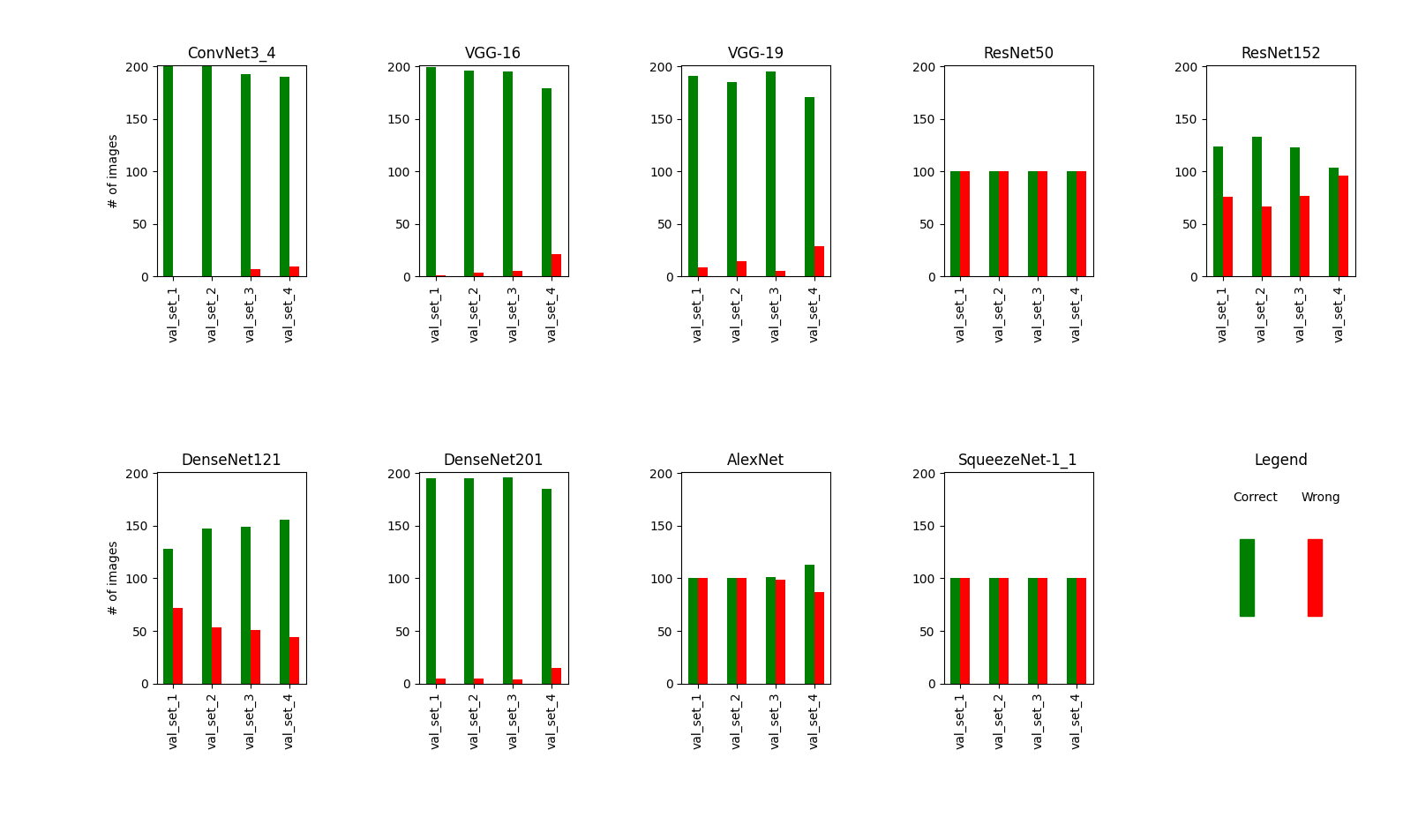}
    \caption{Post-validation bar plots for the $2_{{output-labels}}$ case. As far as the ConvNet3\_4 is concerned, data refer to the best model.}
    \label{fig:validation2}
  \end{center}
\end{figure}

The results of the post-validation phase are shown in Fig.~\ref{fig:validation2} and Table~\ref{tab:validation2}. For each dataset, we computed the validation error of each model. The outcomes show how the ConvNet3\_4 model significantly outperform the other pre-trained models (Kruskal-Wallis H test, $p < 0.05$). In particular, our model obtains the lowest error in 3 out of 4 validation sets (see Table~\ref{tab:validation2}), which is a considerably good result. Regarding the pre-trained models, the VGG-16, VGG-19 and DenseNet-201 networks are competitive with the ConvNet3\_4 model. Conversely, the ResNet-50, ResNet-152, DenseNet-121, AlexNet and SqueezeNet1.1 are remarkably inferior to our model and fail in classifying images of the validation sets ${val\_set}_1$ - ${val\_set}_4$.

\begin{table}[htbp!]
  \begin{center}
    {\caption{Validation error of the models in the $2_{{output-labels}}$ case. As far as the ConvNet3\_4 is concerned, data refer to the best model. Bold values indicate the best results.}\label{tab:validation2}}
    \begin{tabular}{|c|c|c|c|c|}
      \hline
      Model & ${val\_set}_1$ & ${val\_set}_2$ & ${val\_set}_3$ & ${val\_set}_4$ \\
      \hline
      ConvNet3\_4 & \textbf{0.0} & \textbf{0.0} & 0.035 & \textbf{0.05} \\
      \hline
      VGG-16 & 0.005 & 0.02 & 0.025 & 0.105 \\
      \hline
      VGG-19 & 0.045 & 0.075 & 0.025 & 0.145 \\
      \hline
      ResNet-50 & 0.5 & 0.5 & 0.5 & 0.5 \\
      \hline
      ResNet-152 & 0.38 & 0.335 & 0.385 & 0.48 \\
      \hline
      DenseNet-121 & 0.36 & 0.265 & 0.255 & 0.22 \\
      \hline
      DenseNet-201 & 0.025 & 0.025 & \textbf{0.02} & 0.075 \\
      \hline
      AlexNet & 0.5 & 0.5 & 0.495 & 0.435 \\
      \hline
      SqueezeNet1.1 & 0.5 & 0.5 & 0.5 & 0.5 \\
      \hline
    \end{tabular}
  \end{center}
\end{table}

An interesting outcome of this analysis is the discrepancy between categorization accuracy on the test set (results in Table~\ref{tab:accuracy2}) and the capability of classifying images on the validation sets (see Table~\ref{tab:validation2}) of some of the pre-trained deep networks (i.e., ResNet-50, ResNet-152, DenseNet-121 and AlexNet). In particular, these models exhibit poor performances even on the data in ${val\_set}_1$, which correspond to images that are more likely to be acquired in the production process under consideration. This is partially surprising since the list of transformations in ${val\_set}_1$ (Table~\ref{tab:valTrans1}) can be seen as a subset of the transformations in Table~\ref{tab:trainTrans}. A possible explanation might be related to the overfitting issue: minimal changes in the input images make these models generate significant differences in the output predictions. Instead, the other pre-trained networks (i.e., VGG-16, VGG-19 and DenseNet-201) and the ConvNet3\_4 model are more robust to such perturbations.

\subsection{$4_{{output-labels}}$ case}
\label{output4}

In this section we present the results achieved in the most complicated condition, i.e. the $4_{{output-labels}}$ case, in which the model is required to distinguish both the presence or absence of the anticoagulant inside the test tube and the type of vial.

\subsubsection{ConvNet3\_4}
\label{convNet3_4-res4}

\begin{figure}[htbp!]
    \centering
        \begin{subfigure}[b]{0.45\textwidth}
            \centering
            \includegraphics[width=\textwidth]{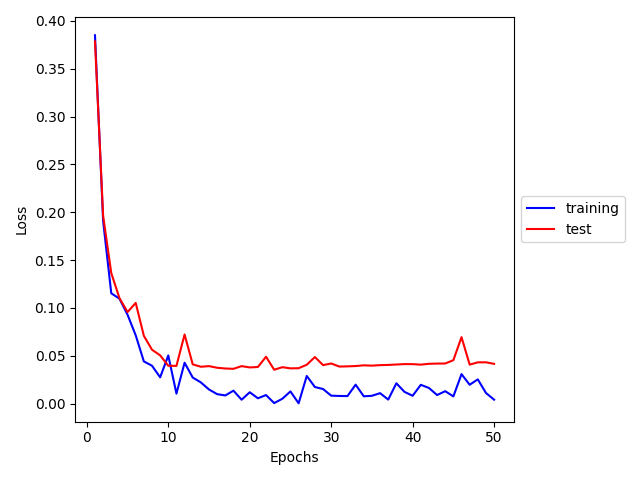}
        \end{subfigure}
        \hfill
        \begin{subfigure}[b]{0.45\textwidth}
            \centering
            \includegraphics[width=\textwidth]{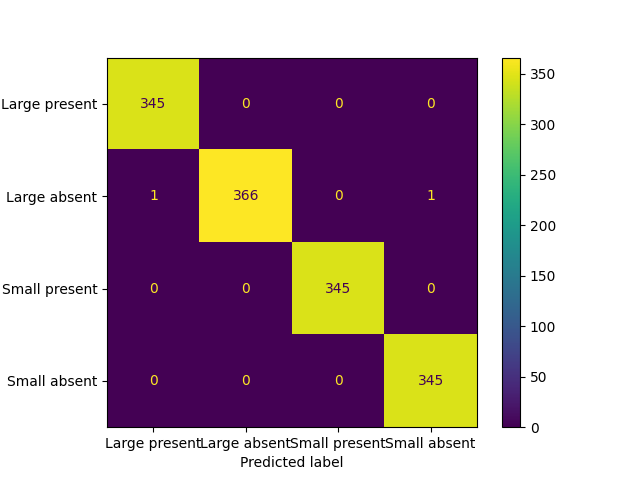}
        \end{subfigure}
    \caption{Results achieved by the ConvNet3\_4 model in the $4_{{output-labels}}$ case. Left: error curve during training Data are obtained by averaging 10 replications of the experiment. Right: confusion matrix. Data refer to the classification capability of our best model with respect to the images in the test set.}
    \label{fig:convNetRes4}
\end{figure}

The ConvNet3\_4 model performed very well also in this challenging scenario. In fact, the average accuracy on the test set is 99.073\%. The error curve of the ConvNet3\_4 model during training and test phases is shown in Fig.~\ref{fig:convNetRes4}, left. The blue curve represents the classification error w.r.t. the images in the training set, while the red curve refers to the test set. As can be seen, the training error is around 0 at the end of the training. Conversely, starting at epoch 10, the test error stabilizes around a value of 0.05. The result is quite good considering the size of the training and test sets. The best model reached an accuracy level of 99.857\%, i.e. a quasi optimal performance. Specifically, it correctly categorized 1401 out of 1403 test images (see Fig.~\ref{fig:convNetRes4}, right), which is a remarkable result given the complexity of the problem.

\begin{table}[htbp!]
  \begin{center}
    {\caption{Performance of the models in the $4_{{output-labels}}$ case. Because of the presence of four possible output classes, in this scenario we computed only the accuracy (\textit{Acc}) metric. As far as the ConvNet3\_4 model is concerned, data refer to the best model. Bold values indicate the best results we obtained.}\label{tab:accuracy4}}
    \begin{tabular}{|c|c|}
      \hline
      \textbf{Model} & \textbf{Acc} \\
      \hline
      ConvNet3\_4 & \textbf{0.9986} \\
      \hline
      VGG-16 & 0.9922 \\
      \hline
      VGG-19 & 0.9964 \\
      \hline
      ResNet-50 & 0.8774 \\
      \hline
      ResNet-152 & 0.881 \\
      \hline
      DenseNet-121 & 0.99 \\
      \hline
      DenseNet-201 & 0.9943 \\
      \hline
      AlexNet & 0.9922 \\
      \hline
      SqueezeNet1.1 & 0.2459 \\
      \hline
    \end{tabular}
  \end{center}
\end{table}

\subsubsection{Pre-trained models}
\label{pre-trained-res4}

Similarly to the $2_{{output-labels}}$ case, we analyzed the accuracy of the pre-trained models in this scenario. Results are provided in Table~\ref{tab:accuracy4}. As can be observed, there is a drop in performance due to the increased complexity of the classification task. This is more evident for the ResNet-50 and ResNet-152 models, whose performance goes below 0.9, and for the SqueezeNet1.1 network, which obtains a very low result of 0.2459 due to its tendency to classify all the images according to one output class.

\begin{figure}[htbp!]
    \begin{center}
        \includegraphics[angle=0,width=\textwidth]{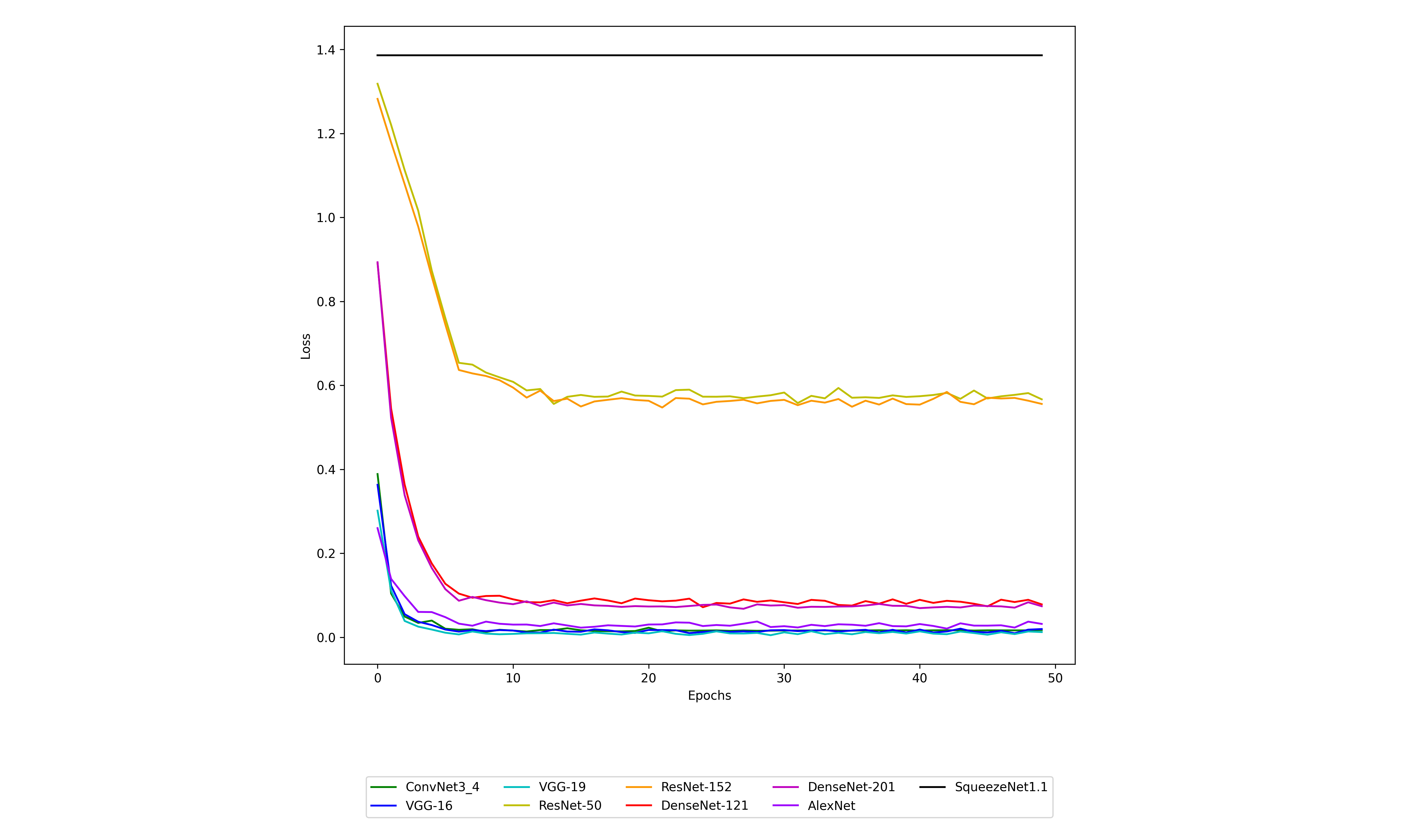}
        \caption{Error curves of the different models in the $4_{{output-labels}}$ case. Data refer to the classification capability on the test set. Concerning the ConvNet3\_4 model, data refer to the best model.}
        \label{fig:loss4}
    \end{center}
\end{figure}

The classification error of the models on the images of the test set is shown in Fig.~\ref{fig:loss4}. In particular, we can observe how the ConvNet3\_4 model and the pre-trained VGG-16, VGG-19 and AlexNet models quickly converge on a quasi-optimal performance (test error around 0.0). The SqueezeNet1.1 model does not improve its classification capability during the training. The ResNet-50 and ResNet-152 models soon reach an error of around 0.6, but they do not manage to further improve their accuracy. Finally, the DenseNet-121 and DenseNet-201 converge to a test error around 0.2.

\begin{table}[htbp!]
  \begin{center}
    {\caption{Validation error of the models in the $4_{{output-labels}}$ case. As far as the ConvNet3\_4 is concerned, data refer to the best model. Bold values indicate the best results.}\label{tab:validation4}}
    \begin{tabular}{|c|c|c|c|c|}
      \hline
      Model & ${val\_set}_1$ & ${val\_set}_2$ & ${val\_set}_3$ & ${val\_set}_4$ \\
      \hline
      ConvNet3\_4 & \textbf{0.01} & \textbf{0.025} & \textbf{0.105} & \textbf{0.07}\\
      \hline
      VGG-16 & 0.5 & 0.51 & 0.5 & 0.55 \\
      \hline
      VGG-19 & 0.515 & 0.505 & 0.5 & 0.575 \\
      \hline
      ResNet-50 & 0.735 & 0.75 & 0.75 & 0.745 \\
      \hline
      ResNet-152 & 0.675 & 0.695 & 0.715 & 0.74 \\
      \hline
      DenseNet-121 & 0.65 & 0.725 & 0.71 & 0.66 \\
      \hline
      DenseNet-201 & 0.145 & 0.17 & 0.205 & 0.15 \\
      \hline
      AlexNet & 0.755 & 0.75 & 0.74 & 0.745 \\
      \hline
      SqueezeNet1.1 & 0.75 & 0.75 & 0.75 & 0.75 \\
      \hline
    \end{tabular}
  \end{center}
\end{table}

\begin{figure}[htbp!]
  \begin{center}
    \includegraphics[angle=0,width=\textwidth]{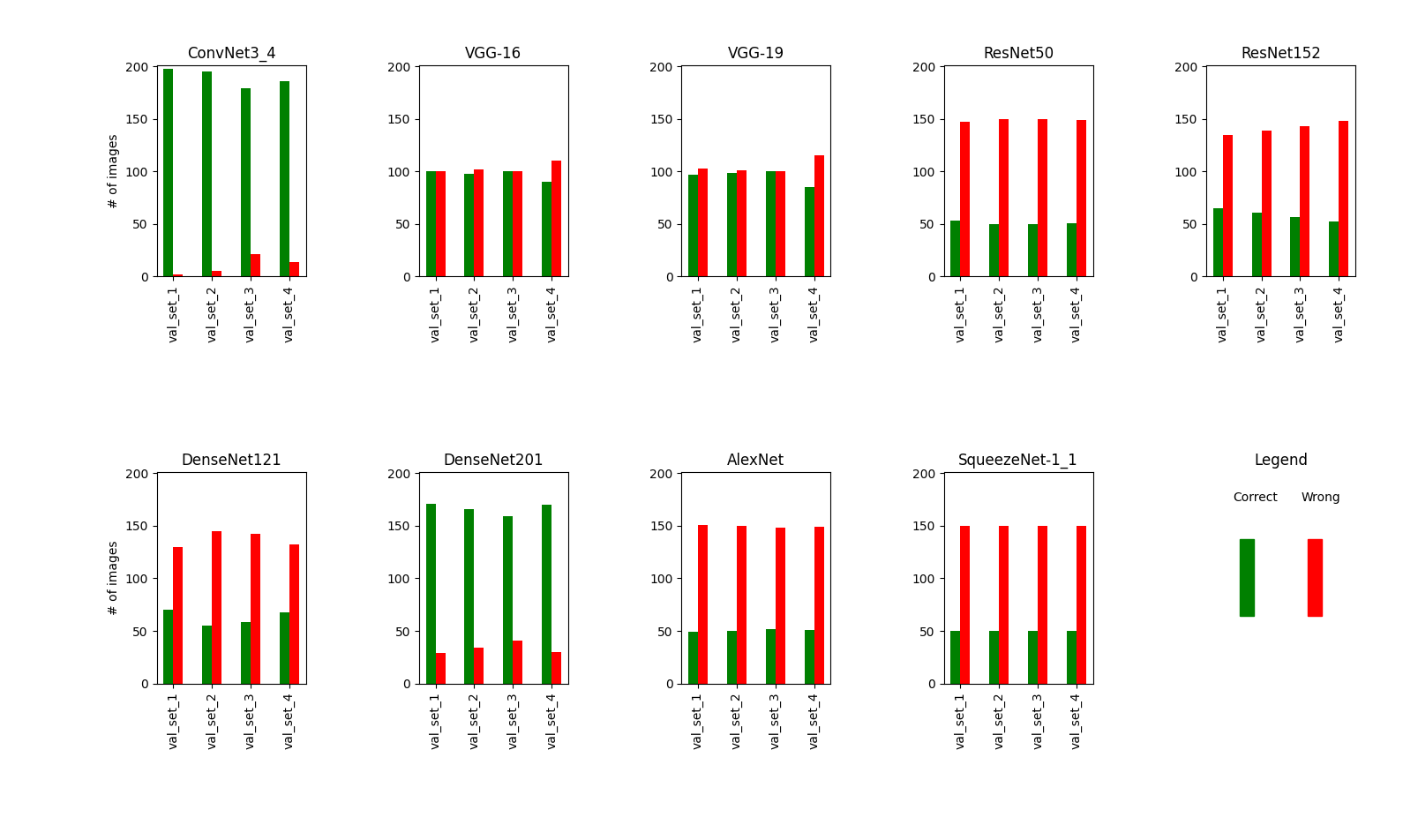}
    \caption{Post-validation bar plots for the $4_{{output-labels}}$ case. As far as the ConvNet3\_4 is concerned, data refer to the best model.}
    \label{fig:validation4}
  \end{center}
\end{figure}

Also in this case, we post-evaluated the capability of the trained models on the validation sets defined in Tables~\ref{tab:valTrans1}-~\ref{tab:valTrans4}. The results are provided in Table~\ref{tab:validation4} and Fig.~\ref{fig:validation4}. As highlighted in Table~\ref{tab:validation4}, there is a remarkable fall in the classification capability of pre-trained networks. In fact, with the exception of DenseNet-201, the categorization error exceeds 0.5, that is these models are not able to correctly predict the output class for more than half of the images in the validation sets ${val\_set}_1$-${val\_set}_4$. Specifically, the ResNet-50, ResNet-152, DenseNet-121, AlexNet and SqueezeNet1.1 models adopt a sub-optimal strategy consisting of associating each input image to one of the possible output classes. The VGG-16 and VGG-19 models implement a similar strategy and return a prediction for the input image by selecting among two output classes. In addition, unlike the previous case, the categorization ability deteriorates immediately when dealing with the images in ${val\_set}_1$. This means that these models do not generalize at all. Overall, the post-evaluation analysis clearly shows how the ConvNet3\_4 model is considerably better than others at generalizing to new unseen data (Kruskal-Wallis H test, $p < 0.05$). 

\subsection{Feature analysis}
\label{features}

\begin{figure}[htbp!]
    \centering
        \begin{subfigure}[b]{0.3\textwidth}
            \centering
            \includegraphics[width=\textwidth]{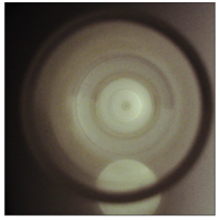}
        \end{subfigure}
        \hfill
        \begin{subfigure}[b]{0.3\textwidth}
            \centering
            \includegraphics[width=\textwidth]{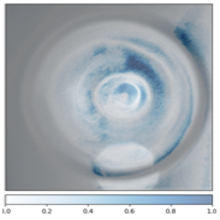}
        \end{subfigure}
        \hfill
        \begin{subfigure}[b]{0.3\textwidth}
            \centering
            \includegraphics[width=\textwidth]{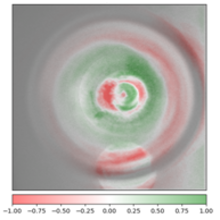}
        \end{subfigure}
    \vfill
    \centering
        \begin{subfigure}[b]{0.3\textwidth}
            \centering
            \includegraphics[width=\textwidth]{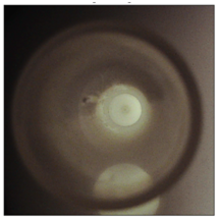}
        \end{subfigure}
        \hfill
        \begin{subfigure}[b]{0.3\textwidth}
            \centering
            \includegraphics[width=\textwidth]{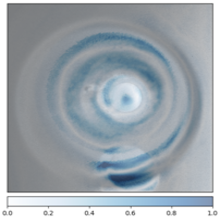}
        \end{subfigure}
        \hfill
        \begin{subfigure}[b]{0.3\textwidth}
            \centering
            \includegraphics[width=\textwidth]{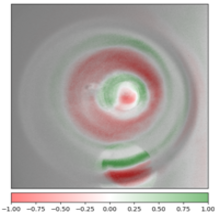}
        \end{subfigure}
    \caption{Comparison between raw image and feature maps. Images refer to the $2_{{output-labels}}$ case. Top: big empty original image (left) and the corresponding \textit{Saliency} (center) and \textit{Integrated Gradients} (right) feature maps. Bottom: big filled original image (left) and the corresponding \textit{Saliency} (center) and \textit{Integrated Gradients} (right) feature maps.}
    \label{fig:featureMaps}
\end{figure}

In this section we provide an analysis of the image features mainly impacting the model's classification. We considered two different techniques for interpreting DL models: \textit{Saliency} \cite{simonyan2013deep} and \textit{Integrated Gradients} \cite{sundararajan2017axiomatic}. ``Saliency'' is a measure of which parts of the image contributed most to the target class prediction, and is represented by the ``Saliency'' map of the input image gradients w.r.t. the specified target class. Null saliency values represent regions of the image having no impact on the target class prediction, while positive saliency values indicate regions having a positive impact on the prediction \cite{simonyan2013deep}. Fig.~\ref{fig:featureMaps} (center) represents the heat map of saliency of the input image shown in Fig.~\ref{fig:featureMaps} (left). The color bar specifies the intensity of saliency, with white representing absence of saliency and blue indicating positive saliency, i.e. blue regions are most influential in deep learning model classification decision.

The ``Integrated Gradients'' heat map represents the integrated attribution of each pixel to the model's classification decision \cite{sundararajan2017axiomatic}. This heat map displays the regions of the image that most influenced the model's classification decision by taking into account the entire input-output trajectory and the reference input distribution (baselines) used in the attribution calculation. The ``Integrated Gradients'' heat map was generated by using the Integrated Gradients technique to calculate the importance of each pixel in the input image for the model prediction \cite{sundararajan2017axiomatic}. Positive values indicate that the corresponding pixels were deemed important. In the ``Integrated Gradients'' heat map shown in Fig.~\ref{fig:featureMaps} (right), colors represent the degree to which each region of the image is predicted by the model. Red corresponds to negative attributions, while green corresponds to positive attributions.

As can be observed, both the ``Saliency'' and the ``Integrated Gradients'' methods indicated how the central part of the input image is paramount for the model classification. However, our analysis reveals also an effect of the white circle in the bottom part of the images (Fig.~\ref{fig:featureMaps}, bottom center and bottom right figures), which corresponds to the space where a second vial could be inserted into the production process.

\section{Conclusions}
\label{conclusions}

In this paper, we present an automated system that is able to monitor the presence or the absence of transparent anticoagulants. The goal is to integrate such a system into the production process of a company that handles plastic consumables to increase efficiency and reduce waste, in line with the Industry 4.0 paradigm. Specifically, we designed a deep network architecture, named ConvNet3\_4, for the ability to detect the presence or the absence of an anticoagulant in test tubes ($2_{{output-labels}}$ case) and we performed a comparison with the state-of-the-art models. In a second experiment, we trained the models to classify not only the presence or the absence of the anticoagulant, but also the size of the vials ($4_{{output-labels}}$ case). The results in Section~\ref{results} show that the proposed model is competitive with the best state-of-the-art models and superior in terms of generalization capability, i.e. the ability to correctly classify images previously not seen during training. This is also true for the more challenging $4_{{output-labels}}$ case. Since industrial monitoring requires a high level of effectiveness, including a high degree of accuracy and the ability to work properly with new data, only the ConvNet3\_4 model demonstrates the performance level required for real industrial applications, while pre-trained networks drop dramatically in performance in the $4_{{output-labels}}$ scenario. However, it is worth noting that the ConvNet3\_4 model has significantly
more parameters than the pre-trained networks, which may improve their
performance. We are planning to carry out a series of experiments in
which we compare the models with the same number of parameters.

In the future, we plan to investigate other techniques to further improve both the classification performance of our model and the time required for training. A first issue we want to address is the minimization of the error fluctuations (see Fig.~\ref{fig:convNetRes4}, left), which may depend on the diversities observed in the input images. One strategy to avoid this type of oscillations is to use adaptive weight decay \cite{ghiasi2022adaptive,nakamura2019adaptive} during training, for example when there is an increase in the error value from one epoch to the next (see epochs 12 and 46 in Fig.~\ref{fig:convNetRes4}, left). 
Furthermore, as we discussed in Section~\ref{features}, the input images contain external components (i.e., the location of a second test tube) that affect the prediction of the model. The actual impact of such behavior has not been investigated in depth, although it is more likely to act as a random noise. Future work should be devoted to such analysis in order to improve the accuracy of the ConvNet3\_4 model and avoid bias. To this end, we are collecting a larger dataset in the laboratory, both to reduce the use of the data augmentation technique (Section~\ref{dataset}) and to better assess the validity of our approach in a real-world monitoring scenario.

\section*{Acknowledgments}
The work of F.P. was partially supported by PNRR-CN1 SPOKE 6 Spoke 6 - Multiscale modeling engineering applications B83C22002940006 under the MUR National Recovery and Resilience Plan funded by the European Union - NextGenerationEU.

\bibliographystyle{unsrt} 

\newpage

\setcounter{table}{0}
\renewcommand{\thetable}{A\arabic{table}}

\appendix

\section{Validation sets}
\label{appendix:val_sets}

\begin{table}[htbp!]
  \begin{center}
    {\caption{List of transformations applied to generate the validation set ${val\_set}_1$}\label{tab:valTrans1}}
    \begin{tabular}{|c|c|c|}
      \hline
      Type & \# of transforms & Parameter \\
      \hline
      Rotation & 5 & random angle in range $[210^\circ,330^\circ]$ \\
      \hline
      Blur & 2 & \begin{tabular}{@{}c@{}} random sigma in range $[0.1,5.0]$ \\ kernel size = 5 \end{tabular} \\
      \hline
      Blur & 2 & \begin{tabular}{@{}c@{}} random sigma in range $[0.1,5.0]$ \\ kernel size = 9 \end{tabular} \\
      \hline
      Posterization & 1 & 6 bits \\
      \hline
    \end{tabular}
  \end{center}
\end{table}

\begin{table}[htbp!]
  \begin{center}
    {\caption{List of transformations applied to generate the validation set ${val\_set}_2$}\label{tab:valTrans2}}
    \begin{tabular}{|c|c|c|}
      \hline
      Type & \# of transforms & Parameter \\
      \hline
      Jitter & 1 & \begin{tabular}{@{}c@{}} brightness = 0.0 \\ contrast = 0.0 \\ saturation = 0.0 \\ hue = 0.0 \end{tabular} \\
      \hline
      \begin{tabular}{@{}c@{}}Rotation +\\ horizontal flipping\end{tabular} & 4 & \begin{tabular}{@{}c@{}}Rotation: \\ random angle in range $[30^\circ,330^\circ]$ \end{tabular} \\
      \hline
      \begin{tabular}{@{}c@{}}Rotation +\\ vertical flipping\end{tabular} & 4 & \begin{tabular}{@{}c@{}}Rotation: \\ random angle in range $[30^\circ,330^\circ]$ \end{tabular} \\
      \hline
      Autocontrast & 1 & \\
      \hline
    \end{tabular}
  \end{center}
\end{table}

\begin{table}[htbp!]
  \begin{center}
    {\caption{List of transformations applied to generate the validation set ${val\_set}_3$}\label{tab:valTrans3}}
    \begin{tabular}{|c|c|c|}
      \hline
      Type & \# of transforms & Parameter \\
      \hline
      Rotation & 3 & random angle in range $[5^\circ,175^\circ]$ \\
      \hline
      Rotation & 3 & random angle in range $[185^\circ,355^\circ]$ \\
      \hline
      \begin{tabular}{@{}c@{}} Center crop + \\ pad \end{tabular} & 1 & \begin{tabular}{@{}c@{}} Center crop: \\ size = 300 \\ Pad: \\ size = 50 \\ fill = 0 \end{tabular} \\
      \hline
      \begin{tabular}{@{}c@{}} Center crop + \\ pad \end{tabular} & 1 & \begin{tabular}{@{}c@{}} Center crop: \\ size = 350 \\ Pad: \\ size = 25 \\ fill = 0 \end{tabular} \\
      \hline
      \begin{tabular}{@{}c@{}} Equalization + \\ horizontal flipping \end{tabular} & 1 & \\
      \hline
      \begin{tabular}{@{}c@{}} Equalization + \\ vertical flipping \end{tabular} & 1 & \\
      \hline
    \end{tabular}
  \end{center}
\end{table}

\begin{table}[htbp!]
  \begin{center}
    {\caption{List of transformations applied to generate the validation set ${val\_set}_4$}\label{tab:valTrans4}}
    \begin{tabular}{|c|c|c|}
      \hline
      Type & \# of transforms & Parameter \\
      \hline
      Sharpness & 1 & sharpness\_factor = 0.5 \\
      \hline
      Sharpness & 1 & sharpness\_factor = 1.5 \\
      \hline
      Sharpness & 1 & sharpness\_factor = 2.5 \\
      \hline
      Sharpness & 1 & sharpness\_factor = 3 \\
      \hline
      Sharpness & 1 & sharpness\_factor = 4 \\
      \hline
      \begin{tabular}{@{}c@{}} Inversion + \\ horizontal flipping \end{tabular} & 1 & \\
      \hline
      \begin{tabular}{@{}c@{}} Inversion + \\ vertical flipping \end{tabular} & 1 & \\
      \hline
      Blur & 3 & \begin{tabular}{@{}c@{}} random sigma in range $[0.1,5.0]$ \\ kernel size = 7 \end{tabular} \\
      \hline
    \end{tabular}
  \end{center}
\end{table}

\end{document}